\documentclass[11pt,a4paper]{article}
\usepackage[hyperref]{acl2017}
\usepackage{times}
\usepackage{latexsym}
\usepackage{graphicx}
\usepackage{comment}
\usepackage{graphics}

\usepackage{url}
\usepackage{tikz}
\usetikzlibrary{shapes.geometric, arrows}

\aclfinalcopy 
\def\aclpaperid{58} 

\title{Enterprise to Computer: Star Trek chatbot}

\author{
Grishma Jena \\\And  Mansi Vashisht \\\And Abheek Basu \\ Computer \& Information Science\\
University of Pennsylvania \\ {\tt gjena, vmansi, abheek, ungar, joao@cis.upenn.edu} \\\And Lyle Ungar \\\And Jo\~ao Sedoc \\}

\date{}

\begin{document}
\maketitle
\begin{abstract}
Human interactions and human-computer interactions are strongly influenced by style as well as content. Adding a persona to a chatbot makes it more human-like and contributes to a better and more engaging user experience. In this work, we propose a design for a chatbot that captures the ‘style’ of Star Trek by incorporating references from the show along with peculiar tones of the fictional characters therein. Our Enterprise to Computer bot (E2Cbot) treats Star Trek dialog style and general dialog style differently, using two recurrent neural network Encoder-Decoder models. The Star Trek dialog style uses sequence to sequence (SEQ2SEQ) models \cite{DBLP:journals/corr/SutskeverVL14, DBLP:journals/corr/BahdanauCB14} trained on Star Trek dialogs. The general dialog style uses Word Graph to shift the response of the SEQ2SEQ model into the Star Trek domain. We evaluate the bot both in terms of perplexity and word overlap with Star Trek vocabulary and subjectively using human evaluators. 
\end{abstract}

\section{Introduction}
Incorporating a specific persona is vital to make chatbots appear more human-like. 
By focusing on the bot's linguistic style, its personality can be extended. We examine various methods to build a chatbot that attempts to capture the characteristic tones of Star Trek characters.  

In this work, we incorporate this aspect of persona by using multiple Encoder-Decoder models to handle different types of dialog. Our Enterprise to Computer bot (E2Cbot) consists of two sequence to sequence (SEQ2SEQ) models \cite{DBLP:journals/corr/SutskeverVL14, DBLP:journals/corr/BahdanauCB14}, one to handle Star Trek style input and another for everyday conversations. We use a Word Graph \cite{filippova2010multi,banerjee2016transforming} trained on Star Trek dialogs for shifting everyday conversation to the Star Trek domain. Sentences are ranked using a Bigram Language Model trained on Star Trek dialogs to find the most likely response. 

The responses of the different models are evaluated using the relative perplexity of the utterance compared to Star Trek dialogs and replaced with alternates if they are too different. This ensures that the chatbot does not give an ungrammatical or incoherent response if the model confidence is below a certain threshold. 

A key challenge with  SEQ2SEQ models is that they tend to frequently give uninteresting responses like "I don't know" or "I don't understand." To ensure an engaging conversation, we define rules to output standard responses when the model confidence is low. Unlike prior research, E2Cbot has independent Encoder-Decoder models for Star Trek and non-Star Trek domains. This structure gives promising results for Star Trek as-is. For non-Star Trek, we insert words into the SEQ2SEQ output to shift it towards the Star Trek domain. 

We foresee a wide variety of applications for models like ours. The resulting chatbot can be used in various social situations as it can mimic different personae. For example, better emotional connection is established when an interlocutor matches the style and interests of the speaker. One could imagine sports-lover bots or extravert bots. 
Similarly, an ideal counselor would be more understanding and would encourage a person to talk whereas a general conversation bot would be more chatty and take a similar number of conversation turns as the user. Also, the utterance length of a counselor would be smaller compared to that of a general conversation bot.

\section{Related work}

\noindent This work is similar in intent to the rule-based ‘Fake Spock’ bot \cite{SpockBot} that imitates Spock. However, since that is a rule-based bot, it becomes too complex to create and update the rules. Fake Spock bot's outputs are illustrated in Table 1. E2Cbot is data-driven and can be trained to encompass multiple topics. It does not imitate any specific Star Trek character, but adds a Star Trek ‘touch’ to its replies about day-to-day topics.   

\begin{table}[h!]
\begin{flushleft}
\resizebox{\columnwidth}{!}{%
\begin{tabular}{|l| l |c|c| } 
 \hline
 User & Fake Spock Bot \\
 \hline \hline
 I am hungry captain. & What will you be eating? \\ 
 You can do it. & I shall attempt it. \\ 
 Mister Spock. & I am an AI chatting on the Internet. \\  
 Maximum power. & I see no logic in preferring Stonn over me.\\

 \hline

\end{tabular}%
}
 \end{flushleft}
\caption{Responses by baseline Pandora bot.}

\end{table}

In the work of \newcite{DBLP:journals/corr/LiGBGD16} personality is embedded into the SEQ2SEQ model to handle speaker inconsistency in response generation. They modified the LSTM cell to encode speaker information and inject it into the hidden layer at each time step. This is called the Speaker Model and models the personality of the speaker and helps in predicting personalized responses throughout the generation process.

Our work is similar to the Neural-Storyteller model \cite{kiros2015skip} which involves 'style-shifting' i.e. transferring standard image captions to the style of stories from novels. Each passage from a novel is mapped to a skip-thought vector. The RNN conditions on the skip-thought vector and generates the passage that it has encoded. It uses a linear vector transformation F(x) to transform input \textit{x} from caption style vector \textit{c} to book-style vector \textit{b} using the equation:

\begin{center}
\textit{F(x) = x - c + b}
\end{center}

\section{Dataset}

We used three different datasets to train individual parts of our model.

\subsection{Star Trek Dataset}
To train the Star Trek SEQ2SEQ component, we created our own dataset of dialogs pulled from various Star Trek T.V. episodes and movies \cite{STdialog}. The initial cleaning was done using an open-source Github repository \cite{STcleaning}. This was followed by rule based cleaning to remove stage directions. Post-response pairs were created using a method similar to that outlined by \newcite{lowe2015ubuntu} for the Ubuntu dialog corpus. The same exchange between characters was used to generate multiple pairs by including the context as well. Exchange $A\rightarrow B\rightarrow C$ gave $(A,B), (B,C)$ and $(AB,C)$ post reply pairs.  The final dataset consisted of 100,990 post-response pairs with an average utterance length of 14.3 words. 

\subsection{Cornell Movie-Dialog Corpus}

We used the Cornell Movie-Dialogue Corpus by \newcite{Danescu-Niculescu-Mizil+Lee:11a} to train a SEQ2SEQ component to handle general, non-Star Trek conversations. It contained 199,455 post-response pairs with an average utterance length of 12.82 words.

\subsection{Tweets}

We used an open source Twitter dataset \cite{Twitter} to train a binary classifier to better predict non-Star Trek style inputs. This dataset is meant to capture regular, non-Star Trek conversation that a user might attempt to have with E2Cbot. We used 50,000 post response tweet pairs with an average utterance length of 16.18 words.

\section{E2Cbot}

\noindent Figure 1 shows the pipeline of our model\footnote{Our code is available at \url{https://github.com/GJena/CIS-700-7_Chatbot-Project}}. We use multiple SEQ2SEQ models to cover Star Trek-like dialogs and normal dialogs. To handle everyday conversation dialogs, we used the Cornell movie dataset. We will discuss the model in detail in the following subsections.

\tikzstyle{startstop} = [rectangle, rounded corners, minimum width=2cm, minimum height=1cm,text centered, draw=black]
\tikzstyle{io} = [diamond, minimum width=3cm, minimum height=1cm, text centered, draw=white]
\tikzstyle{arrow} = [thick,->,>=stealth]

\begin{figure*}[ht!]

\resizebox{\textwidth}{!}{%

\begin{tikzpicture}[node distance=2cm]

\node (ip) [io] {User Utterance};
\node (Clf) [startstop, , right of=ip,  xshift = 2 cm] {Binary Classifier};
\node (SEQ2SEQ_ST) [startstop, right of=Clf,  yshift=2cm, xshift = 2 cm] {Star Trek SEQ2SEQ};
\node (SEQ2SEQ_NST) [startstop, right of=Clf,  yshift=-2cm, xshift =2 cm] {Cornell Movie Data SEQ2SEQ};
\node (WG) [startstop, right of=SEQ2SEQ_NST, xshift = 3cm] {Word Graph};
\node (LM) [startstop, right of=WG, xshift = 2cm] {Language Model};
\node (op) [io, right of=LM, xshift = 2cm, yshift = 2cm] {Response};

\draw [arrow] (ip) -- (Clf);
\draw [arrow] (Clf) -- (SEQ2SEQ_ST);
\draw [arrow] (Clf) -- (SEQ2SEQ_NST);
\draw [arrow] (SEQ2SEQ_NST) -- (WG);
\draw [arrow] (WG) -- (LM);
\draw [arrow] (LM) -| (op);
\draw [arrow] (SEQ2SEQ_ST) -| (op);

\end{tikzpicture}%
}
\caption{Star Trek Bot Pipeline}
\end{figure*}
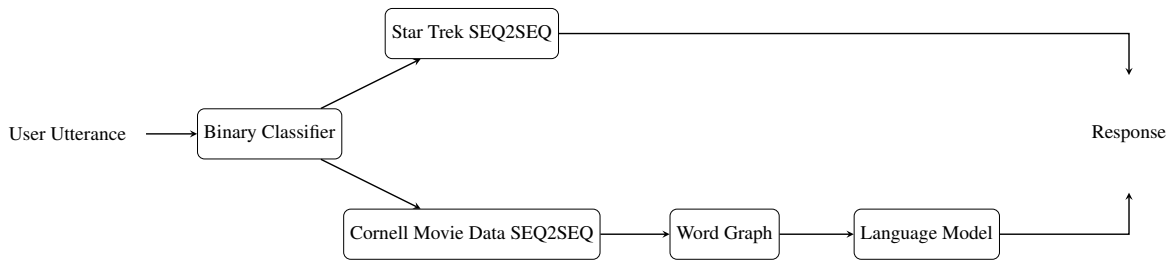

\subsection{Binary Classifier}
A logistic regression-based binary classifier routes the user utterance to either the Star Trek SEQ2SEQ model or the Cornell Movie Data SEQ2SEQ model. The classifier was trained on 200,000 Star Trek dialogs, 100,000 Cornell Movie Dialog Corpus dialogs and 100,000 tweets from the Twitter dataset.  From them, we randomly sampled 80\% dialogs as training and 20\% dialogs as test data. The feature space was constructed using top 10,000 term frequency-inverse document frequency (TF-IDF) unigrams and bigrams after removing  stop words. The classifier had a 95\% accuracy on the test set.

\subsection{Star Trek SEQ2SEQ}
The first SEQ2SEQ model was trained on Star Trek data. It has 3 hidden layers with 1024 units encoder decoder structure. To compensate for the lack of data, we augmented the data by adding context as mentioned in Section 3.1.

\subsection{Cornell Movie Data SEQ2SEQ}
The second SEQ2SEQ model was trained on Cornell Movie data. Its architecture is the same as the Star Trek SEQ2SEQ model. 

\subsection{Word Graph}

We used a modified implementation of the Word Graph algorithm by \citet{banerjee2016transforming} for domain-specific linguistic styles. Star Trek dialogs were selected to construct a word graph that stores words and their POS tags as nodes and adjacency as edges. The output generated by the normal conversation SEQ2SEQ model is tokenized and parsed through the NLTK POS tagger \cite{Loper:2002:NNL:1118108.1118117} i.e. each node represents (\textless word\textgreater, \textless POS\textgreater). Words with multiple POS tags are added as different nodes. The algorithm parses through the input and looks up the word graph for a list of candidate words that can be added between any two words in the input or at the start or end of the input.

This implementation is especially good at inserting words such as ‘Doctor’, ‘Jim’ at the start or end of the sentences due to high occurrence in the Star Trek dialogs. A few examples of Word Graph output are shown in Figure 2.

\begin{figure}[ht!]

\noindent \texttt{I am sorry \textbf{Miranda} \newline}
\noindent \texttt{I will go \textbf{back} \newline}
\noindent \texttt{\textbf{Uhura} how are you\newline}
\noindent \texttt{\textbf{Captain} shall I leave}
\caption{Generated responses for normal conversation SEQ2SEQ. The words in bold have been added by the Word Graph algorithm.}

\end{figure}

Apart from adding names of characters to the sentence, the Word Graph algorithm can also append other words to make sentences grammatical. Figure 3 shows relevant examples.

\begin{figure}[ht!]

\noindent \texttt{This I think it's not \textbf{true} \newline}
\noindent \texttt{\textbf{Feeling} fine. \newline}

\caption{Generated responses for normal conversation SEQ2SEQ. The words in bold have been added by the Word Graph algorithm to make sentences grammatical.}

\end{figure}

\subsection{Filtering using a Language Model}
Since the Word Graph produces some ungrammatical outputs, a Bigram Language Model trained on the augmented Star Trek dialog corpus is used. The sentence with the highest probability is chosen. If multiple sentences have the highest probability, we choose the one containing words present in a handcrafted keyword list.

\subsection{Filtering Unlikely Response Candidates}
If the perplexity of the response is very low or very high compared to the perplexity of Star Trek dialogs, a response from a standard response set or a reply in Klingon is output.

Both  filtering techniques disposed of ungrammatical and non-Star Trek response candidates to ensure high quality output.

\section{Evaluation}
There is little consensus on the best evaluation metrics for free-form chatbots. We used a set of standard input sentences against which we evaluate both bots. 

Our evaluation dataset consists of 20 sentences of which 50\% are normal conversation and 50\% are Star Trek specific dialogs. Quantitative metrics used include perplexity, overlap with Star Trek vocabulary and human evaluation. Perplexity of the response is compared with the perplexity of Star Trek dialogs. Overlap with the Star Trek vocabulary is measured with the rationale that a higher overlap would better capture the Star Trek style.

In addition, ten human annotators rate the responses on the properties of correct grammar, coherence or relevance and Star Trek relatedness. The annotators give a score of 0 if the response does not exhibit the property or 1 if it does. The scorers comprised six people who are Star Trek fans and four who aren't familiar with Star Trek.

\section{Results}
Figure 4 shows some sample responses of the bots for our evaluation dataset. Both output valid responses, except for "Engage". "Engage" is a command generally given to activate the warp drive of the spaceship. E2Cbot gives a coherent response whereas Fake Spock bot's reply is irrelevant. This shows the shortcoming of pure rule based system, since it is difficult to cover all cases.  In the last example, E2Cbot adds \textit{Spock} to the sentence to give it a Star Trek touch but the Pandora bot gives a generic response. Table 2 shows the scores given by the annotators for different metrics. Table 3 shows perplexity of responses on the Star Trek dialog data and the vocabulary overlap of the responses with the Star Trek vocabulary. The perplexity of Star Trek dialogs was found out to be 65.69.

\begin{table}[h!]
\begin{center}

\begin{tabular}{ |c|c|c| } 
 \hline
 Metric & Pandora Bot & E2Cbot \\
 \hline \hline
 Grammar & \textbf{98\%} & 93.5\% \\ 
 Coherence & 60.5\% & \textbf{73.5\%} \\ 
 Star Trek style & 64\% & \textbf{86\%} \\
 \hline
 Average score & 74.16\% & \textbf{84.33\%} \\
 \hline
\end{tabular}
\caption{The human evaluations to compare the Pandora bot with the E2Cbot model.}
\end{center}
\end{table}

\begin{table}[th!]
\begin{center}
\resizebox{\columnwidth}{!}{%
\begin{tabular}{ |c|c|c| } 
 \hline
 Model & Average Perplexity & Vocabulary Overlap\\
 \hline \hline
 Pandora Bot & 45.77 & 95.31\%\\
 E2Cbot & 60.92 & 91.45\%\\
 \hline
\end{tabular}%
}
\caption{Average perplexity and vocabulary overlap of various models on the Star Trek dataset.}
\end{center}
\end{table}

The overall performance of E2Cbot is better than the Fake Spock Pandora bot. However, the Pandora bot does a better job of generating grammatically correct responses since it is rule-based. Our data-driven model is able to produce more coherent responses, including responding to out of domain input. Additionally, the responses generated by E2Cbot had more Star Trek style.

\begin{figure}[h!]
\textbf{User:} Red alert. \newline
\textbf{E2Cbot:} Captain Uhura what's happening. \newline
\textbf{Fake Spock:} Acknowledged. Taking emergency action.
\vspace{0.25em}
\hrule
\vspace{0.25em}
\textbf{User:} Engage. \newline
\textbf{E2Cbot:} Warp one sir. \newline
\textbf{Fake Spock:} I am a Vulcan.
\vspace{0.25em}
\hrule
\vspace{0.25em}
\textbf{User:} Do you like me. \newline
\textbf{E2Cbot:} Bones I like you. \newline
\textbf{Fake Spock:} Sure, Lieutenant I like you very much.
\vspace{0.25em}
\hrule
\vspace{0.25em}
\textbf{User:} Shall I leave? \newline
\textbf{E2Cbot:} No  Spock. \newline
\textbf{Fake Spock:} Not right now, maybe later.
\caption{Sample responses of E2Cbot and Pandora bot.}
\end{figure}

\section{Conclusion}
Our model is able to automatically generate text in Star Trek style, even for out of domain input. It is, in general, an important advance beyond rule based systems like Fake Spock bot. Since we are mainly using a data driven approach, this model can be easily expanded to other domains like news or sports. It can also be extended to emulate specific fictional characters. Further exploration can be done to combine the two models and achieve a superior model. 

Future work involves experimenting with MemN2N by \citet{DBLP:journals/corr/SukhbaatarSWF15} and Joint Attention mechanism by \citet{DBLP:journals/corr/XingWWLHZM16} in place of SEQ2SEQ. MemN2N has shown to retain information for a long period of time. Joint Attention mechanism allows the encoder to focus on multiple things. Additionally the model might be augmented by explicitly adding a personality vector along with tone and mood as input.

\bibliography{acl2017}
\bibliographystyle{acl_natbib}

\end{document}